%% file: main.tex
\pdfoutput=1

\documentclass[11pt]{article}

\usepackage[final]{acl}

\usepackage{times}
\usepackage{latexsym}

\usepackage[T1]{fontenc}

\usepackage[utf8]{inputenc}

\usepackage{microtype}

\usepackage{inconsolata}

\input{own_command}

\title{
Harnessing Large Language Models as Post-hoc Correctors
}

\author{
    Zhiqiang Zhong \\
    Aarhus University \\
    \texttt{zzhong@cs.au.dk} \\
    \And
    Kuangyu Zhou \\
    Microsoft \\
    \texttt{luckyjoou@gmail.com} \\
    \And
    Davide Mottin \\
    Aarhus University \\
    \texttt{davide@cs.au.dk} \\
}

\begin{document}
\maketitle

\begin{abstract} 
\input{pages/abstract}
\end{abstract}

\section{Introduction} 
\label{sec:introduction}
\input{pages/introduction}

\section{Related Work} 
\label{sec:related_work}
\input{pages/related_work}

\section{Preliminary and Problem} 
\label{sec:preliminary}
\input{pages/preliminary}

\section{Methodology} 
\label{sec:framework}
\input{pages/framework}

\section{Are LLMs Post-hoc Correctors?} 
\label{sec:experimental_study}
\input{pages/experiments}

\section{Concluding Discussion} 
\label{sec:conclusion}
\input{pages/conclusion}

\input{pages/statement}

\section*{Acknowledgments}
This work is supported by the Horizon Europe and Innovation Fund Denmark under the Eureka, Eurostar grant no E115712 - AAVanguard.

\bibliography{full_format_references}

\newpage
\appendix
\input{pages/appendix}

\end{document}

%% file: own_command.tex
\usepackage{adjustbox}
\usepackage{makecell}
\usepackage{multirow}
\usepackage[ruled,vlined,linesnumbered]{algorithm2e}
\usepackage{enumitem}
\usepackage{amsmath,amsthm,amsfonts}
\usepackage{xcolor}
\usepackage{xspace}
\usepackage{mdframed}
\newmdtheoremenv[%
  backgroundcolor=gray!20,
  linecolor=red!60!black,
  linewidth=2pt,
  topline=false,
  rightline=false,
  skipabove=10pt,
  skipbelow=10pt,
  leftline=false]{regbox}{Box}

\newcommand{\specialcell}[2][c]{%
    \begin{tabular}[#1]{@{}c@{}}#2\end{tabular}
}

\newcommand{\lipo}{\texttt{ogbg-mollipo}\xspace}
\newcommand{\freesolv}{\texttt{ogbg-molfreesolv}\xspace}
\newcommand{\esol}{\texttt{ogbg-molesol}\xspace}
\newcommand{\bbbp}{\texttt{ogbg-molbbbp}\xspace}
\newcommand{\bace}{\texttt{ogbg-molbace}\xspace}
\newcommand{\hiv}{\texttt{ogbg-molhiv}\xspace}
\newcommand{\twitter}{\texttt{Twitter}\xspace}
\newcommand{\cora}{\texttt{Cora}\xspace}

\newcommand{\IP}{\textbf{IP}\xspace}
\newcommand{\IPD}{\textbf{IPD}\xspace}
\newcommand{\IE}{\textbf{IE}\xspace}
\newcommand{\IED}{\textbf{IED}\xspace}
\newcommand{\FS}{\textbf{FS}\xspace}

\newcommand{\FSD}{\textbf{FSD}\xspace}

\newcommand{\eg}{\emph{e.g.}}
\newcommand{\ie}{\emph{i.e.}}

\newcommand{\model}{\textsc{LlmCorr}\xspace}


%% file: pages/abstract.tex
As Machine Learning (ML) models grow in size and demand higher-quality training data, the expenses associated with re-training and fine-tuning these models are escalating rapidly. 
Inspired by recent impressive achievements of Large Language Models (LLMs) in different fields, this paper delves into the question: \emph{can LLMs efficiently improve an ML's performance at a minimal cost?} 
We show that, through our proposed training-free framework \model, an LLM can work as a post-hoc \emph{corrector} to propose \emph{corrections} for the predictions of an arbitrary ML model. 
In particular, we form a contextual knowledge database by incorporating the dataset's label information and the ML model's predictions on the validation dataset. 
Leveraging the in-context learning capability of LLMs, we ask the LLM to summarise the instances in which the ML model makes mistakes and the correlation between primary predictions and true labels. 
Following this, the LLM can transfer its acquired knowledge to suggest corrections for the ML model's predictions. 
Our experimental results on text analysis and the challenging molecular predictions show that \model improves the performance of a number of models by up to $39\%$
\footnote{The code and models are available at \url{https://github.com/zhiqiangzhongddu/LLMCorr}.}. 

%% file: pages/introduction.tex
In recent decades, Machine Learning (ML) models have become increasingly prevalent in various real-world applications~\cite{DHB20,ZBM23}. 
As ML models grow in size and demand higher-quality training data, the expenses associated with pre-training and fine-tuning these models to achieve superior performances are escalating rapidly~\cite{DCLT18,HGC21}. 
Hence, there is an urgent need to develop effective, lightweight and practical solutions for users to improve their ML model's predictions. 

\begin{figure}[!t]
\centering
\includegraphics[width=.8\linewidth]{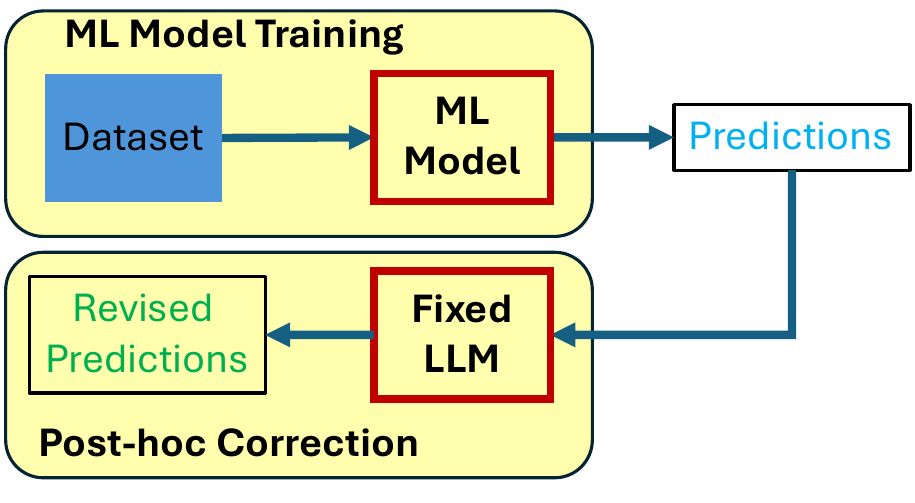}
\caption{
Harnessing LLMs as post-hoc \emph{correctors.}
A fixed LLM is leveraged to propose corrections to an arbitrary ML model's predictions without additional training or the need for additional datasets. 
}
\label{fig:concept}
\end{figure}

Large Language Models (LLMs) exhibit unprecedented capabilities in understanding and generating human-like text, making them invaluable across a wide range of Natural Language Processing (NLP) tasks, including machine translation~\cite{AMAV23}, commonsense reasoning~\cite{KMSG23} and coding tasks~\cite{BCEG23}. 
While LLMs have showcased their effectiveness across an array of NLP applications, the full extent of their potential in broader fields remains largely unexplored~\cite{ZDLW24}. 
This paper delves into the essential research question: \emph{can LLMs effectively improve an ML's performance at a minimal cost?}

To answer the research question, we propose a groundbreaking framework, \model, which extends the application scope of LLMs by positioning them as training-free post-hoc \emph{correctors} (illustrated in Figure~\ref{fig:concept}). 
A \emph{fixed} LLM is leveraged to propose corrections to an arbitrary ML model's predictions without introducing additional training or the need for additional datasets. 

At its core, \model consists of three main steps: (1) After completing the training of the ML model, we collect the dataset's available label information, along with the primary predictions made by the ML model on the validation set, to construct a contextual knowledge database.
We anticipate that the established database records the contextual knowledge about the types of instances for which the ML models produce accurate or inaccurate predictions, as well as the correlation between the primary predictions and the true labels. 
A recent breakthrough known as In-Context Learning (ICL) \cite{LYFJ23} has enhanced the adaptability of LLMs by enabling them to acquire contextual knowledge during inference, eliminating the need for extensive fine-tuning \cite{CTR20}. 
(2) Consequently, given the target data with a primary prediction generated by the ML model, we extract the relevant contextual knowledge from the knowledge database to form a prompt. 
Given the input token constraints of current LLMs, transmitting all contextual information to one prompt for querying becomes impractical \cite{TMSA23,AAAA23}. 
To address this limitation, we introduce an embedding-based information retrieval approach, which can efficiently locate similar data likely to offer relevant insights with the target data.
(3) Finally, we query the LLM using the created prompts for suggestions to refine the target data's primary prediction.
To mitigate the knowledge hallucination~\cite {HYMZ23}, we implement a \emph{self-correction} mechanism when the LLM demonstrates a tendency to make substantial modifications to the ML model's prediction. 

The training-free nature of \model carries several natural advantages: \textit{(i)} \model facilitates the straightforward application of LLMs, eliminating the necessity for expensive re-training and fine-tuning for an arbitrary ML model. 
\textit{(ii)} One of the most obvious limitations of LLMs is their reliance on \emph{unstructured} text~\cite{GDL23}, but \model can be adapted to arbitrary scenarios by incorporating different ML models. 

To validate the effectiveness of \model, we mainly deploy it in the face of structured molecule graph data within the domain of biology, \eg, predicting the functionality of molecules. 
Through extensive experiments conducted on six real-world benchmark datasets covering diverse subjects, we empirically demonstrate that \model significantly elevates the quality of predictions across diverse ML models, achieving notable improvements, up to 39\%.
Besides, we demonstrate \model's effectiveness on general text analysis tasks.
Furthermore, we conduct comprehensive follow-up ablation studies and analyses to validate the efficacy of \model's designs and elucidate the impact of key factors. 

%% file: pages/related_work.tex
\textbf{Large Language Models}.
Traditional language models are typically trained on sequences of tokens, learning the likelihood of the next token dependent on the previous tokens~\cite{VSPUJGKP17}. 
Recently, \citet{BMRS20} demonstrated that increasing the size of language models and the amount of training data can result in new capabilities, such as zero-shot generalisation, where models can perform text-based tasks without specific task-oriented training data. 
Consequently, Large Language Models (LLMs) have experienced exponential growth in both size and capability in recent years~\cite{BMRS20,TMSA23,AAAA23}. 
A wide range of NLP applications have been reshaped by LLMs, including machine translation~\cite{AMAV23}, commonsense reasoning~\cite{KMSG23} and coding tasks~\cite{BCEG23}.
In this work, we innovatively harness LLMs as post-hoc \textit{correctors}, further extending the application scope of LLMs. 

\smallskip
\noindent
\textbf{In-Context Learning}.
While the impressive performance and generalisation capabilities of language models have rendered them highly effective across various tasks~\cite{WTBR22}, they have also resulted in larger model parameters and increased computational costs for additional fine-tuning on new downstream tasks~\cite{HSWA21}. 
To address this challenge, recent research has introduced In-Context Learning (ICL), enabling LLMs to excel at new tasks by incorporating a few task samples directly into the prompt~\cite{LYFJ23}. 
Despite the success of these methods in improving LLM performance, their potential for correcting predictions made by ML models has not been thoroughly explored.
This work investigates the utility of LLMs as post-hoc \emph{correctors} to rectify incorrect predictions by leveraging their ICL abilities.

\smallskip
\noindent
\textbf{Post-hoc Corrector for Machine Learning Models.}
Driven by the increasing prevalence of ML models in diverse real-world applications ~\cite{BDCIW18,DHB20,ZBM23}, academia and industry have invested significant efforts in enhancing ML effectiveness. 
However, the majority of existing research focuses on refining the design of the ML module.
Meanwhile, as ML models grow in size and demand higher-quality training data, the expenses associated with re-training and fine-tuning these models are escalating rapidly~\cite{DCLT18,HGC21}. 
Hence, there is an urgent need to develop effective, lightweight and practical solutions for users to improve their ML model's predictions adaptively. 
While some studies~\cite{HHSLB21,ZIP22} propose post-processing techniques to adjust ML model predictions on node-wise tasks of graph-structured data, their solutions often lack scalability across different types of data.
This work introduces a novel and versatile post-hoc \textit{corrector} framework applicable to an arbitrary ML model. 

%% file: pages/preliminary.tex
This paper aims to leverage LLMs as post-hoc \emph{correctors} to enhance predictions made by an arbitrary ML model. 
Specifically, we showcase our framework on challenging prediction tasks on structured molecule graph data in biology. 
We further demonstrate the generality of our framework on sentiment analysis and text classification in Section~\label{sec:appendix_broader_tasks} in the Appendix.
For instance, consider the supervised molecule property prediction task, where molecules can be represented using various formats such as \emph{SMILES string}~\cite{W88} and \emph{geometry structures}~\cite{ZDLW24} (as shown in Figure~\ref{fig:molecule_info}). 
However, a notable limitation of existing LLMs is their reliance on unstructured text, rendering them unable to incorporate essential geometry structures as input~\cite{LLWL23,GDL23}. 
To overcome this limitation, \citet{FHP23} propose encoding the graph structure into text descriptions. 
In this paper, as depicted in Figure~\ref{fig:molecule_info}, 
we extend this concept by encoding both the molecule's atom features and graph structure into textual \emph{descriptions}.

\smallskip
\noindent
\textbf{Notion.}
Given a molecule, we formally represent it as a graph $\mathcal{G} = (S, G, D)$, where $S$, $G$ and $D$ denote the \emph{SMILES string}, \emph{geometry structures} and generated atom feature and graph structure \emph{descriptions} of $\mathcal{G}$. 
$y \in \mathcal{Y}$ stands for the label for $\mathcal{G}$. 

\smallskip
\noindent
\textbf{Problem Setup.}
Given a set of molecules $\mathcal{M} = \{ \mathcal{G}_{1}, \mathcal{G}_{2}, \dots, \mathcal{G}_{m}\}$, where $\mathcal{M}_{\mathcal{T}} \subset \mathcal{M}$ contains molecules with known labels $y_{v}$ for all $\mathcal{G}_{v} \in \mathcal{M}_{\mathcal{T}}$. 
Our objective is to predict the unknown labels $y_{u}$ for all $\mathcal{G}_{u} \in \mathcal{M}_{test}$, where $ \mathcal{M}_{test} = \mathcal{M} \setminus \mathcal{M}_{\mathcal{T}}$. 
In addition, $\mathcal{M}_{\mathcal{T}}$ is split into two subsets: $\mathcal{M}_{train}$ and $\mathcal{M}_{val}$, where $\mathcal{M}_{train}$ is the training set and $\mathcal{M}_{val}$ works as the validation set. 

\smallskip
\noindent
\textbf{ML Models.}
The conventional approach to tackle molecule property prediction tasks is employing ML models. 
Take the supervised molecule property prediction task as an example. 
The goal is to learn a mapping function $f_{ML}: \mathcal{M} \to \hat{\mathcal{Y}}$, by minimising loss function value $\min_{\Theta} \sum_{i=1}^{n} \mathcal{L}(\hat{\mathcal{Y}}_{train}^{i}, \mathcal{Y}_{train}^{i})$, where $\Theta$ represents the set of trainable parameters of $f_{ML}$. 
Subsequently, $f_{ML}$ can be employed on test dataset $\mathcal{M}_{test}$ to generate predictions $\hat{\mathcal{Y}}_{test}$. 

\smallskip
\noindent
\textbf{Leveraging LLMs as Post-hoc \emph{Correctors}.}
In recent decades, significant efforts have been devoted to enhancing the effectiveness, robustness, and generalisation of advanced ML models ($f_{ML}$). 
However, the potential for improving the quality of ML model predictions after completing the training process remains largely unexplored.
With the trend of ML models increasing in size and requiring higher-quality training data, the costs associated with re-training and fine-tuning ML models are rapidly escalating. 
This paper intends to explore the possibility of positioning LLMs ($f_{LLM}$) as post-hoc \emph{correctors} to refine predictions of an arbitrary $f_{ML}$. 
Compared with re-training and fine-tuning a model, this work has outstanding advantages in terms of cost and versatility. 

%% file: pages/framework.tex
\begin{figure*}[!ht]
\centering
\includegraphics[width=.9\linewidth]{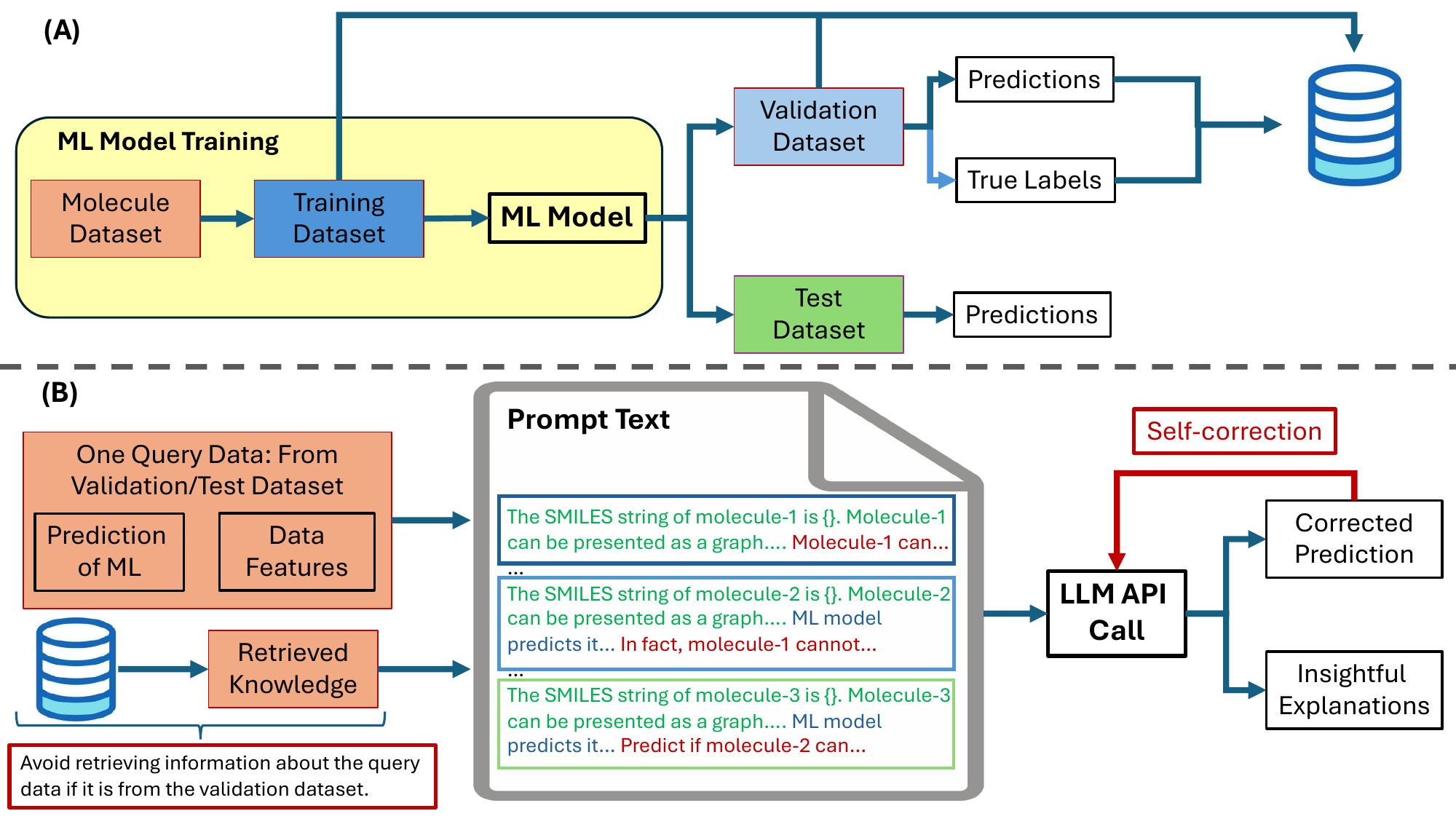}
\caption{
A high-level overview of \model, harnessing Large Language Models (LLMs) as post-hoc \textit{correctors} to refine predictions made by an arbitrary Machine Learning (ML) model. 
}
\label{fig:framework}
\end{figure*}

In this section, we outline the workflow of \model, designed as a post-hoc \textit{corrector} to refine predictions generated by any ML model. 
As illustrated in Figure~\ref{fig:framework}, the key idea is to leverage the LLM's ICL ability to summarise the types of data in which the ML model makes mistakes and the correlation between primary predictions and true labels, thereby refining the ML's prediction on the test dataset. 
To achieve this goal, \model comprises three main steps: (1) Contextual knowledge database construction; (2) Contextual knowledge retrieval; (3) Prompt engineering and query. 

\subsection{Contextual Knowledge Database Construction}
\label{subsec:context_knowledge_database}
After completing the training of the ML model ($f_{ML}$) on the training set $\mathcal{M}_{train}$, we collect data from both the training set $\mathcal{M}_{train}$ and the validation set $\mathcal{M}_{val}$, along with the primary predictions $\hat{\mathcal{Y}}_{val}$ made by $f_{ML}$ on the validation set.
Subsequently, we construct a contextual knowledge database $\mathcal{D}$ based on the collected data. 
Notably, the knowledge database $\mathcal{D}$ not only contains information regarding the original training and validation datasets but also provides insights into the types of molecules for which the ML models generate accurate or inaccurate predictions.
Additionally, it captures the relationship between the initial prediction $\hat{\mathcal{Y}}_{val}$ and the true label $\mathcal{Y}_{val}$. 
We anticipate that this essential contextual knowledge will empower the LLM ($f_{LLM}$) to refine the predictions $\hat{\mathcal{Y}}_{test}$ made by $f_{ML}$ on the test dataset $\mathcal{M}_{test}$.

\subsection{Contextual Knowledge Retrieval}
\label{subsec:knowledge_retrieval}
The effectiveness of the \model heavily relies on the richness and relevance of the information received by the LLM, as it determines the task-specific contextual knowledge available to the LLM. 
However, due to the input token constraints of existing LLMs, transmitting all contextual knowledge into the LLM is impractical \cite{TMSA23,AAAA23}.
After constructing the contextual knowledge database $\mathcal{D}$, the next challenge is to retrieve relevant contextual knowledge for a given query data $\mathcal{Q}_u = (\mathcal{G}_u, \hat{y}_u)$ from either the validation set $\mathcal{M}_{val}$ or the test set $\mathcal{M}_{test}$. 
To address this limitation, we propose an \emph{Embedding-based Information Retrieval} (EIR) approach.
The EIR technique comprises two primary steps: (1) Utilising a text encoder ($f_{Emb}$) on available textual descriptions of molecule $\mathcal{G}_u$, we generate embedding vectors $f_{Emb}: (S, D) \to \mathbf{Z}$ for molecules from the knowledge database $\mathcal{D}$ and the query data $\mathcal{Q}_u$.
(2) Calculating the similarity between the query data $\mathbf{Z}_u$ and molecules in the knowledge database based on the obtained embeddings $\mathbf{Z}_v, \forall \mathcal{G}_v \in \mathcal{D}$. 
Different selection strategies can be employed to retrieve various contextual knowledge based on the ranking. 
In this study, we retrieve the top-$k$ similar data as the contextual knowledge.
We delve into the influence of different retrieval selection strategies in Section~\ref{subsec:ablation_study}.

\smallskip
\noindent
\textbf{Addressing Data Leakage Concerns.}
It is important to recognise that when the query data $\mathcal{Q}_u$ comes from the validation dataset $\mathcal{M}_{val}$, precautions are taken to prevent retrieving information about $\mathcal{Q}_u$ to avoid data leakage. 
Then, \model's performance on $\mathcal{M}_{val}$ can be assessed in a manner consistent with traditional ML pipeline, \eg, hyper-parameters selection. 

\subsection{\model Prompt Engineering}
\label{subsec:prompt_engineering}

\begin{figure}[!ht]
\centering
\includegraphics[width=1.\linewidth]{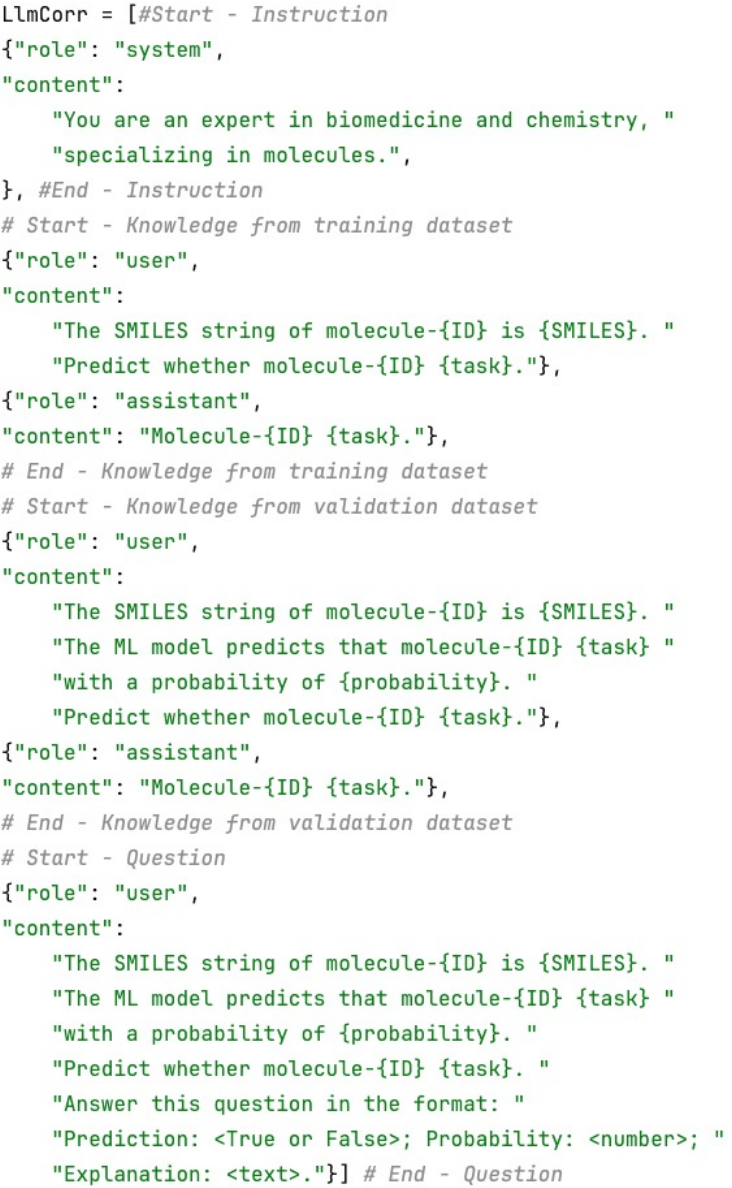}
\caption{
\model prompt template. 
Multiple contextual knowledge from training and validation datasets can be included by expanding the template. 
}
\label{fig:prompt_corrector}
\end{figure}

The goal in prompt engineering is to find the correct way to formulate a question $\mathcal{Q}$ in such a way that an LLM ($f_{LLM}$) will return the corresponding answer $A$ essentially represented as $A = f_{LLM}(\mathcal{Q})$. 
In this work, our goal is to provide the LLM with helpful and comprehensive contextual knowledge regarding molecules and the ML model's behaviours on the validation set so that it can make precise corrections to the ML model's predictions on the test dataset. 
A variety of approaches exist for modifying the $f_{LLM}$ so that it could better perform downstream tasks such as fine-tuning~\cite{CTR20} and LoRA~\cite{HSWA21}. 
However, these methods typically require access to the internals of the model and heavy computation capability, which can limit their applicability in many real-world scenarios. 
In this work, we are instead interested in the case where $f_{LLM}$ and its parameters are fixed, and the system is available only for users in a black box setup where $f_{LLM}$ only consumes and produces text. 
We believe this setting to be particularly valuable as the number of proprietary models available and their hardware demands increase. 

\smallskip
\noindent
\textbf{\model Prompting}.
To this end, we propose a novel prompt that serves to position LLMs as post-hoc \textit{correctors}, refining predictions made by an arbitrary ML model.
Different from existing prompts that often work as \textit{predictors}~\cite{FHP23} or \textit{explainers}~\cite{HBLPLH23}, leveraging LLMs as \textit{correctors} combining the strengths of LLMs' in context-based question answering with the ML model's proficiency in learning from specific datasets.
Specifically, the proposed \model prompt template is illustrated in Figure~\ref{fig:prompt_corrector}, which mainly consists of following components:
\begin{enumerate}[leftmargin=*]\itemsep0em
    \item \textit{Instruction}: Provides general guidance to the LLM, clarifying its role in the conversation.
    \item \textit{Contextual knowledge from the training dataset}: Includes SMILES string and molecule label information of the training dataset. 
    \item \textit{Contextual knowledge from the validation dataset}: Includes SMILES string, molecule label information and the ML model's predictions of the validation dataset. 
    This equips the LLM with insights into the ML model's error patterns, enhancing its ability to refine predictions on the test dataset. 
    \item \textit{Question}: Tasks the LLM to refine the ML model's predictions for the query data, drawing on the provided contextual knowledge.
\end{enumerate}
It's worth noting that by expanding the template's sections on contextual knowledge from the training and validation datasets, multiple instances of contextual knowledge from the knowledge database $\mathcal{D}$ can be included.
Subsequently, we query the LLM with generated prompts to obtain an initial response concerning the refined prediction of the query data, along with probability values and explanations, offering significant interpretability.


\begin{figure}[!ht]
\centering
\includegraphics[width=1.\linewidth]{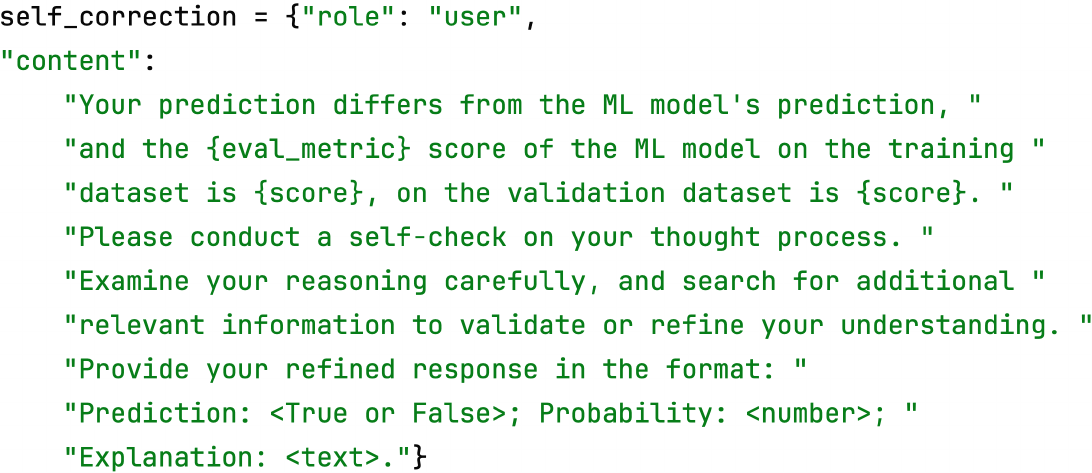}
\caption{
\textit{Self-correction} prompt template. 
}
\label{fig:prompt_self_correction}
\end{figure}

\smallskip\noindent
\textbf{Self-correction Prompting.}
An inherent limitation of existing LLMs is their tendency to generate \textit{hallucinations} producing content that deviates from real-world facts or user inputs~\cite{GASL23,HYMZ23}. 
One promising solution is known as \textit{self-correction}, where the LLM is prompted or guided to rectify errors in its own output~\cite{PSXNWW23}. 
In this work, after obtaining corrected prediction $\Tilde{y}_u$ from the LLM, we introduce a \textit{self-correction} mechanism (prompt template is shown in Figure~\ref{fig:prompt_self_correction}) if the LLM makes significant modifications on the primary prediction generated by the ML (for classification tasks - reversing the prediction label; for regression tasks - modifying the prediction value range by more than 20\%). 
As demonstrated empirically in our experiments (see Section~\ref{subsec:ablation_study}), this approach can prevent LLM from hallucinating incorrect corrections in many cases. 

\smallskip
\noindent
\textbf{Extensibility.}
\model is designed as a post-hoc \textit{corrector} framework that leverages the ICL capability of LLMs. 
As illustrated in Figure~\ref{fig:framework}, \model is adaptable to an arbitrary ML model, showcasing its remarkable extensibility. 
With the trend of ML models increasing in size and requiring higher-quality training data, the costs associated with enhancing ML models by re-training and fine-tuning are rapidly escalating. 
\model emerges as a promising, general-purpose practical solution in the LLM era. 
\model is summarised in Algorithm~\ref{alg:framework} in Appendix~\ref{sec:appendix_algorithm}. 

%% file: pages/experiments.tex
In this section, we evaluate the effectiveness of LLMs as post-hoc \textit{correctors}. 
Our experimental analysis mainly focuses on the challenging structured molecular graph property prediction tasks. 
The additional experiments on general text analysis tasks are presented in Appendix~\ref{sec:appendix_broader_tasks}.

\smallskip\noindent
\textbf{Dataset.}
We consider six widely used molecule datasets from the OGB benchmark \cite{HFRNDL21}, including \bace, \bbbp, \hiv, \esol, \freesolv and \lipo. 
Detailed descriptions are summarised in Appendix~\ref{sec:appendix_dataset}. 

\smallskip\noindent
\textbf{ML Models.}
To investigate whether \model can effectively improve predictions across various types of ML models.
We consider ML models of three different categories: (1) Language Model (LM) that only takes text information as inputs, \ie, DeBERTa~\cite{HGC21}. 
(2) Graph Neural Networks (GNNs) that capture the molecule's geometry structure information. 
We consider two classic GNN variants, \ie, GCN~\cite{KW17} and GIN~\cite{XHLJ19}, and two SOTA GNN variants collected from the OGB leaderboards \cite{HFRNDL21}, \ie, HIG and PAS~\cite{WZYH21}. 
(3) And we consider one recently released hybrid framework, TAPE~\cite{HBLPLH23}, in our experiments. 
TAPE leverages LM and LLMs to capture textual information as features, which can be used to boost GNN performance. 
The implementation details are discussed in Appendix~\ref{sec:appendix_implementation}. 

\smallskip\noindent
\textbf{LLMs.}
In this work, we are interested in where the LLM's parameters are fixed, and the system is available for users in a black box setup where the LLM only consumes and produces text. 
We believe this setting to be particularly valuable as most users would practically have access to LLMs. 
In this case, we consider GPT-3.5 and GPT-4 \cite{AAAA23} as LLMs in this work, and GPT-3.5 is the major LLM for most experiments. 
Besides, open-source LLMs are also of great interest since they are easier to deploy privately, so we adopt Llama2~\cite{TMSA23} as another LLM. 
Because the generated \emph{descriptions} following \cite{FHP23} have tons of tokens, easily over the LLM's input token constraints, hence we do not include descriptions in the \model prompt in this study. 
Including the description information is a promising future work to explore with LLMs of more input tokens. 

\subsection{Main Results}
\label{subsec:main_results}

\input{tables/table-corrector-performance}

\textbf{Observation 1: \model is a potent post-hoc corrector.}
Examining the molecule graph property prediction performance across six datasets in Table~\ref{table:corrector_performance}, it's evident that \model consistently delivers substantial enhancements over various ML models, with improvements reaching up to $39\%$ in terms of RMSE.
This consistent and notable performance underscores the effectiveness of LLMs within our framework \model, serving as proficient post-hoc \emph{correctors} to refine the primary predictions generated by ML models.

\smallskip\noindent
\textbf{Observation 2: The significance of geometric structure.}
Table~\ref{table:corrector_performance} underscores the superiority of models incorporating geometric structure over others.
This highlights the crucial role of geometric structure in accurately predicting a molecule's property.
However, \model currently cannot directly incorporate geometric structure in the prompt due to limitations in the token count of generated descriptions over the LLM's constraints.
Addressing this limitation is identified as a promising avenue for future exploration.

\smallskip\noindent
\textbf{Observation 3: Enhanced assistance for lower-performing ML models.}
Furthermore, we observe that \textit{\model provides more substantial assistance when the performance of ML models is lower.}
This trend is noticeable across various datasets; for instance, \model boosts LM performance from $0.6163$ to $0.6915$ with a $12.2\%$ improvement on the test dataset of \bace.
Even though the ultimate performance still falls short compared to GNN models, the magnitude of improvement is most significant.

\input{tables/table-different-llm}

\smallskip\noindent
\textbf{Observation 4:}
\textbf{GPT-4 and Llama2 underperform compared to GPT-3.5.}
Table~\ref{table:influence_llms} displays the molecule graph property prediction performance and execution time for three datasets, comparing Llama2, GPT-3.5 and GPT-4.
Llama2 has fewer parameters than GPT-3.5 and GPT-4, but Llama2 can still improve the predictions within the framework of \model on some datasets. 
Interestingly, we find that despite its larger training data and more complex fine-tuning process, GPT-4 exhibits inferior performance compared to GPT-3.5 in this study.
We hypothesise that this discrepancy may be attributed to interventions such as reinforcement learning through human feedback.

\input{tables/table-predictor-performance}
\smallskip\noindent
\textbf{Observation 5:}
\textbf{LLMs exhibit limited competitiveness as predictors.}
Given \model's remarkable performance as \emph{correctors}, another intriguing question arises: \emph{can LLM generate accurate predictions directly?}
To investigate, we conduct additional experiments where the LLM is tasked with directly predicting the molecule's property.
For detailed findings, please refer to Appendix~\ref{sec:appendix_additional_experiments} due to space constraints.
As shown in the results of Table~\ref{table:predictor_performance}, LLMs do not demonstrate competitive performance as predictors.
This observation reinforces the efficacy of \model, which leverages LLMs as post-hoc \emph{correctors}.

\subsection{Ablation Study}
\label{subsec:ablation_study}


\input{tables/table-knowledge-select}

\smallskip\noindent
\textbf{Variants of contextual knowledge retrieval.}
Within the EIR of \model, the selection of top-$k$ data from the knowledge database following similarity calculations is a critical step.
This ablation study explores alternative approaches such as \emph{Jump} and \emph{Random}.
In \emph{Random}, $k$ data are randomly selected from the knowledge database, disregarding similarity rankings.
On the other hand, \emph{Jump} selects $k$ evenly spaced indices, ensuring diversity in the selected data.
Results from Table~\ref{table:influence_top_jump_random} suggest that selecting top-$k$ data yields optimal results, with \emph{Jump} outperforming \emph{Random}.
We posit that LLMs benefit from closely relevant knowledge to generate effective corrections.

\begin{figure}[!ht]
\centering
\includegraphics[width=1.\linewidth]{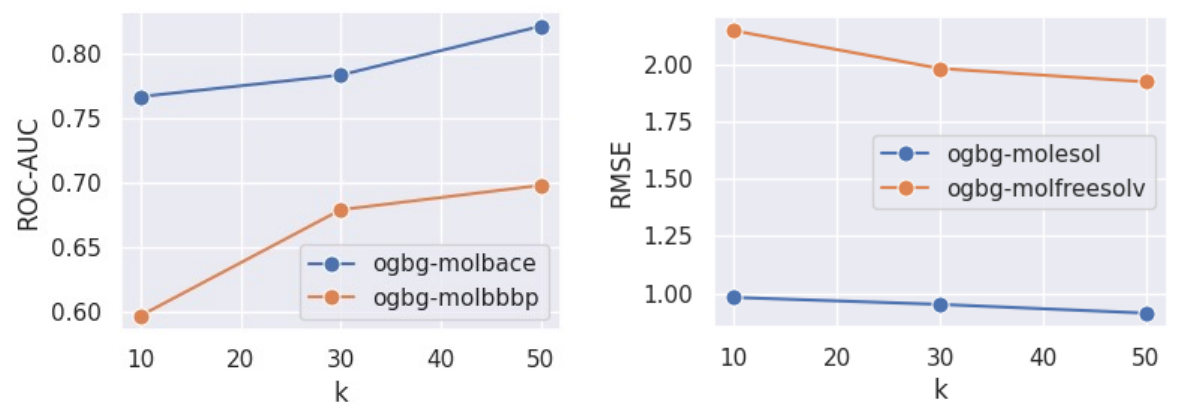}
\caption{
Ablation study of \model with a different number of contextual knowledge data ($k$) on \bace and \esol datasets. 
}
\label{fig:influence_k}
\end{figure}

\smallskip\noindent
\textbf{Impact of $k$.}
In the EIR, the parameter $k$ dictates the number of knowledge instances sampled from the database to construct \model's prompt, thus influencing the knowledge presented to the LLM.
It is observed (Figure~\ref{fig:influence_k}) that larger $k$ values correlate with improved performance, underscoring the significance of comprehensive knowledge to guide LLMs for enhanced performance.

\begin{figure}[!ht]
\centering
\includegraphics[width=1.\linewidth]{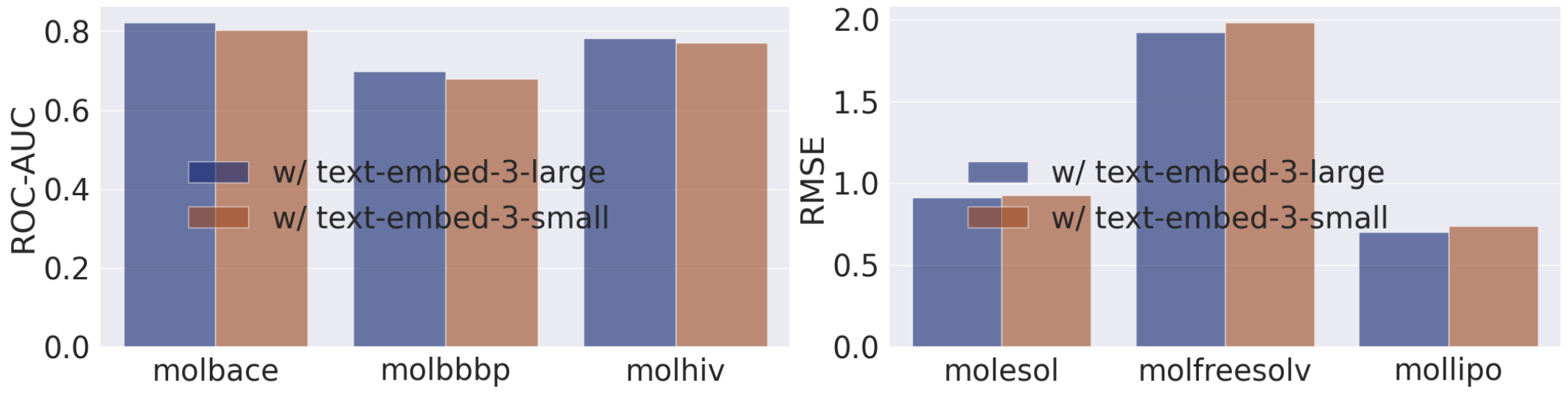}
\caption{
Ablation study of \model on six datasets with different $f_{Emb}$. 
}
\label{fig:influence_f_embd}
\end{figure}

\smallskip\noindent
\textbf{Effect of $f_{Emb}$.}
Another ablation study concerning the EIR process examines the influence of different $f_{Emb}$ functions.
Figure~\ref{fig:influence_f_embd} suggests that larger $f_{Emb}$ values yield superior performance on benchmark testing, aiding \model in achieving better results.
This is attributed to accurate semantic embeddings facilitating the identification of relevant instances during the EIR process, reinforcing the importance of selecting top-$k$ relevant knowledge instances.

\begin{figure}[!ht]
\centering
\includegraphics[width=1.\linewidth]{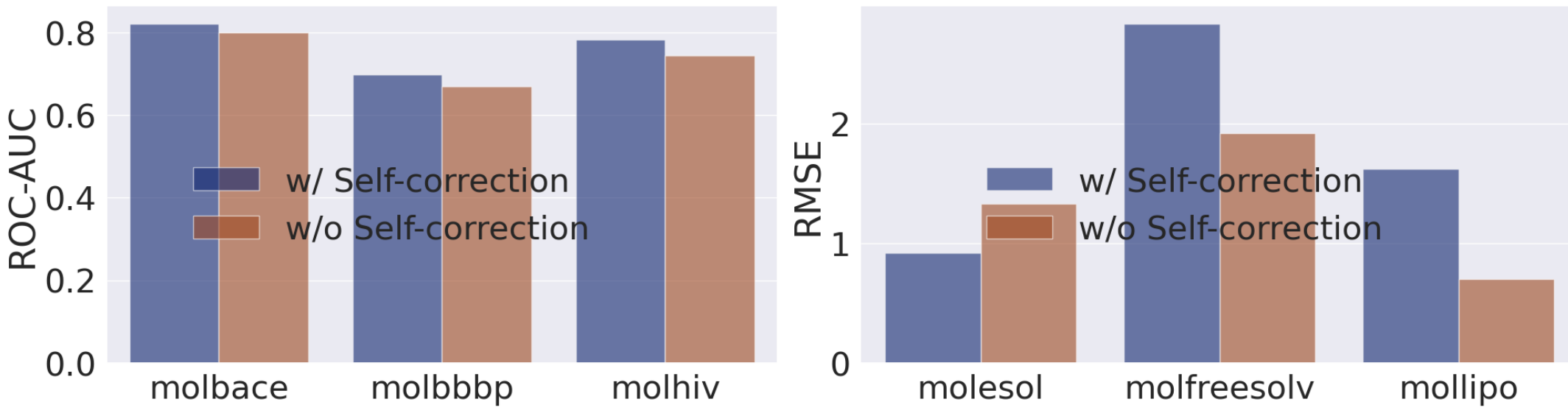}
\caption{
Ablation study of \model on six datasets w/ and w/o self-correction. 
}
\label{fig:self_correction}
\end{figure}

\smallskip\noindent
\textbf{Impact of \emph{self-correction}.}
Upon completion of \model's inference process, the LLM is tasked with self-correction if major modifications to primary predictions are made.
Figure~\ref{fig:self_correction} illustrates \model's performance on the test dataset across six datasets, revealing instances where the self-correction component leads to uncertain impacts.
This phenomenon is attributed to the LLM becoming hesitant and cautious after the questioning. 
Designing a more effective self-correction prompt emerges as an intriguing area for future investigation.

%% file: tables/table-corrector-performance.tex
\begin{table*}[!ht]
\caption{
    Molecule graph property prediction performance for the \bace, \bbbp, \hiv, \esol, \freesolv and \lipo datasets. 
    Classification tasks are evaluated on ROC-AUC ($\uparrow$: higher is better), and regression tasks are evaluated on RMSE ($\downarrow$: lower is better). 
    The improvements of \model over the ML predictive models are reported below \model's performance. 
}
\label{table:corrector_performance}
\centering
\resizebox{1.\linewidth}{!}{
\begin{tabular}{l|cc|cc|cc|cc|cc|cc}
\hline
\hline
 & \multicolumn{2}{c|}{\textbf{\bace}} & \multicolumn{2}{c|}{\textbf{\bbbp}} & \multicolumn{2}{c|}{\textbf{\hiv}} 
 & \multicolumn{2}{c|}{\textbf{\esol}} & \multicolumn{2}{c|}{\textbf{\freesolv}} & \multicolumn{2}{c}{\textbf{\lipo}} \\
 & \multicolumn{6}{c|}{ROC-AUC $\uparrow$} & \multicolumn{6}{c}{RMSE $\downarrow$} \\
\hline
 & Valid & Test & Valid & Test & Valid & Test & Valid & Test & Valid & Test & Valid & Test \\
\hline
\hline
LM              & 0.5584 & 0.6163
                & 0.9307 & 0.6727
                & 0.5024 & 0.5037
                & 2.1139 & 2.2549
                & 6.6189 & 4.4532
                & 1.2095 & 1.1066 \\
LM$^{\model}$    & \specialcell{0.6110\\+9.4\%} & \specialcell{0.6915\\+12.2\%}
                & \specialcell{0.9481\\+1.9\%} & \specialcell{0.6897\\+2.5\%}
                & \specialcell{0.6253\\+24.5\%} & \specialcell{0.6154\\+22.2\%}
                & \specialcell{1.4113\\-33.2\%} & \specialcell{1.3747\\-39.0\%}
                & \specialcell{5.7195\\-13.6\%} & \specialcell{3.5595\\-20.1\%} 
                & \specialcell{1.0210\\-15.6\%} & \specialcell{0.9468\\-14.4\%} \\
\hline
GCN             & 0.7879 & 0.7147
                & 0.9582 & 0.6707
                & 0.8461 & 0.7376
                & 0.8538 & 1.0567
                & 2.8275 & 2.5096
                & 0.6985 & 0.7201 \\
GCN$^{\model}$   & \specialcell{0.8203\\+4.1\%} & \specialcell{0.7718\\+8.0\%}
                & \specialcell{0.9595\\+0.0\%} & \specialcell{0.7045\\+5.0\%}
                & \specialcell{0.8540\\+0.9\%} & \specialcell{0.7529\\+2.1\%}
                & \specialcell{0.7744\\-9.3\%} & \specialcell{0.9108\\-13.8\%}
                & \specialcell{2.0325\\-28.1\%} & \specialcell{2.2102\\-11.9\%} 
                & \specialcell{0.6874\\-1.6\%} & \specialcell{0.7043\\-2.2\%} \\
\hline
GIN             & 0.8042 & 0.7833
                & 0.9611 & 0.6821
                & 0.8406 & 0.7601
                & 0.7685 & 0.9836
                & 2.4141 & 2.2435
                & 0.6503 & 0.7100 \\
GIN$^\model$     & \specialcell{0.8336\\+3.7\%} & \specialcell{0.8214\\+4.9\%}
                & \specialcell{0.9710\\+1.0\%} & \specialcell{0.6982\\+2.4\%}
                & \specialcell{0.8523\\+1.4\%} & \specialcell{0.7822\\+2.9\%}
                & \specialcell{0.7418\\-3.5\%} & \specialcell{0.9137\\-7.1\%}
                & \specialcell{2.1790\\-9.7\%} & \specialcell{1.9219\\-14.3\%} 
                & \specialcell{0.6219\\-4.4\%} & \specialcell{0.6995\\-1.5\%} \\
\hline
TAPE            & 0.7824 & 0.7410
                & 0.9421 & 0.6994
                & 0.8364 & 0.7514
                & 0.8351 & 0.9872
                & 2.8453 & 2.2134
                & 0.6839 & 0.7168 \\
TAPE$^{\model}$  & \specialcell{0.8074\\+3.2\%} & \specialcell{0.7788\\+5.1\%}
                & \specialcell{0.9653\\+2.5\%} & \specialcell{0.6996\\+0.0\%}
                & \specialcell{0.8406\\+0.5\%} & \specialcell{0.7693\\+2.4\%}
                & \specialcell{0.7966\\-4.6\%} & \specialcell{0.9605\\-2.7\%}
                & \specialcell{2.6184\\-8.0\%} & \specialcell{2.0470\\--7.5\%}
                & \specialcell{0.6751\\-1.3\%} & \specialcell{0.7074\\-1.3\%} \\
\hline
HIG             & 0.8213 & 0.8094
                & 0.9730 & 0.6974
                & 0.8400 & 0.8393
                & 0.7756 & 0.9504
                & 2.3590 & 2.2546
                & 0.6130 & 0.7036 \\
HIG$^\model$     & \specialcell{0.8294\\+1.0\%} & \specialcell{0.8135\\+0.5\%}
                & \specialcell{0.9748\\+0.2\%} & \specialcell{0.7074\\+1.4\%}
                & \specialcell{0.8489\\+1.1\%} & \specialcell{0.8447\\+0.6\%}
                & \specialcell{0.7536\\-2.8\%} & \specialcell{0.9322\\-1.9\%}
                & \specialcell{2.3556\\-0.1\%} & \specialcell{1.8799\\-16.6\%}
                & \specialcell{0.6040\\-1.5\%} & \specialcell{0.6920\\-1.6\%} \\ 
\hline
PAS             & 0.8199 & 0.7473
                & 0.9403 & 0.6618
                & 0.8273 & 0.8402
                & 0.8791 & 1.0348
                & 2.3500 & 2.3546
                & 0.6715 & 0.7088 \\
PAS$^\model$     & \specialcell{0.8230\\+0.4\%} & \specialcell{0.7920\\+6.0\%}
                & \specialcell{0.9671\\+2.9\%} & \specialcell{0.6842\\+3.4\%}
                & \specialcell{0.8422\\+1.8\%} & \specialcell{0.8490\\+1.0\%}
                & \specialcell{0.8251\\-6.1\%} & \specialcell{0.9859\\-4.7\%}
                & \specialcell{2.1130\\-10.1\%} & \specialcell{1.9320\\-17.9\%}
                & \specialcell{0.6342\\-5.6\%} & \specialcell{0.6897\\-2.7\%} \\ 
\hline
\end{tabular}
}
\end{table*}

%% file: tables/table-different-llm.tex
\begin{table*}[!ht]
\caption{
    Molecule graph property prediction performance and execution time for the \bace, \esol and \freesolv datasets, with different LLMs. 
    Classification tasks are evaluated on ROC-AUC ($\uparrow$: higher is better), and regression tasks are evaluated on RMSE ($\downarrow$: lower is better). 
}
\label{table:influence_llms}
\centering
\small
\begin{tabular}{l|ccc|ccc|ccc}
\hline
\hline
 & \multicolumn{3}{c|}{\textbf{\bace}}
 & \multicolumn{3}{c|}{\textbf{\esol}} & \multicolumn{3}{c}{\textbf{\freesolv}} \\
 & \multicolumn{2}{c}{ROC-AUC $\uparrow$} & Execution & \multicolumn{2}{c}{RMSE $\downarrow$} & Execution & \multicolumn{2}{c}{RMSE $\downarrow$} & Execution \\
\hline
 & Valid & Test & & Valid & Test & & Valid & Test & \\
\hline
GCN$^\model_{Llama2}$  & 0.7897 & 0.7425 & $\sim$ 14.2 min
                        & 0.9217 & 1.3998 & $\sim$ 16.1 min
                        & 7.2022 & 6.4409 & $\sim$ 17.0 min \\
GCN$^\model_{GPT3.5}$   & 0.8203 & 0.7718 & $\sim$9.5 min
                        & 0.7744 & 0.9108 & $\sim$11.5 min
                        & 2.0325 & 2.2102 & $\sim$10.5 min \\
GCN$^\model_{GPT4}$     & 0.7910 & 0.7713 & $\sim$155 min
                        & 0.8953 & 1.0105 & $\sim$204min
                        & 6.5331 & 3.5777 & $\sim$107min \\
\hline
GIN$^\model_{Llama2}$  & 0.7927 & 0.7745 & $\sim$ 14.5 min
                        & 1.4430 & 1.5921 & $\sim$ 19.3 min
                        & 7.9854 & 7.8273 & $\sim$ 15.6 min \\
GIN$^\model_{GPT3.5}$   & 0.8336 & 0.8214 & $\sim$9.6 min
                        & 0.7418 & 0.9137 & $\sim$12.1 min
                        & 2.1790 & 1.9219 & $\sim$11.7 min \\
GIN$^\model_{GPT4}$     & 0.8022 & 0.7875 & $\sim$148 min
                        & 1.1384 & 0.9552 & $\sim$192min
                        & 7.4731 & 3.9611 & $\sim$112min \\
\hline
\end{tabular}
\end{table*}

%% file: tables/table-predictor-performance.tex
\begin{table*}[!ht]
\caption{
    Molecule graph property prediction performance for the \bace, \bbbp, \hiv, \esol, \freesolv and \lipo datasets. 
    Classification tasks are evaluated on ROC-AUC ($\uparrow$: higher is better), and regression tasks are evaluated on RMSE ($\downarrow$: lower is better). 
}
\label{table:predictor_performance}
\centering
\resizebox{1.\linewidth}{!}{
\begin{tabular}{l|cc|cc|cc|cc|cc|cc}
\hline
\hline
 & \multicolumn{2}{c|}{\textbf{\bace}} & \multicolumn{2}{c|}{\textbf{\bbbp}} & \multicolumn{2}{c|}{\textbf{\hiv}} 
 & \multicolumn{2}{c|}{\textbf{\esol}} & \multicolumn{2}{c|}{\textbf{\freesolv}} & \multicolumn{2}{c}{\textbf{\lipo}} \\
 & \multicolumn{6}{c|}{ROC-AUC $\uparrow$} & \multicolumn{6}{c}{RMSE $\downarrow$} \\
\hline
 & Valid & Test & Valid & Test & Valid & Test & Valid & Test & Valid & Test & Valid & Test \\
\hline
LLM$_{\IP}$     & 0.5690 & 0.5756
                & 0.4606 & 0.5399
                & 0.5519 & 0.5892
                & 2.6221 & 2.0422
                & 6.1699 & 4.4421
                & 1.9836 & 1.8411 \\
LLM$_{\IPD}$    & 0.4835 & 0.5534
                & 0.4643 & 0.4664
                & 0.4732 & 0.5693
                & 3.7395 & 3.1721
                & 8.1598 & 7.2877
                & 2.6464 & 2.5046 \\
LLM$_{\IE}$     & 0.4769 & 0.5220
                & 0.4463 & 0.5237
                & 0.5487 & 0.5419
                & 2.1055 & 2.5549
                & 5.9059 & 4.3097
                & 2.1044 & 1.9158 \\
LLM$_{\IED}$    & 0.5299 & 0.4761
                & 0.4742 & 0.4091
                & 0.5361 & 0.5512
                & 3.9001 & 4.2289
                & 7.4837 & 5.3689
                & 2.4191 & 2.4219 \\
LLM$_{\FS-1}$   & 0.4822 & 0.5122
                & 0.5955 & 0.4954
                & 0.5229 & 0.5268
                & 1.7699 & 2.8762
                & 6.4785 & 4.7553
                & 1.9810 & 1.8432 \\
LLM$_{\FS-2}$   & 0.4277 & 0.6090
                & 0.6019 & 0.5075
                & 0.5619 & 0.5731
                & 1.9271 & 2.1020
                & 5.5078 & 4.5606
                & 1.9138 & 1.8118 \\
LLM$_{\FS-3}$   & 0.5405 & 0.5949
                & 0.6000 & 0.5388
                & 0.5475 & 0.5616
                & 1.9548 & 1.9963
                & 6.3753 & 4.7241
                & 1.8291 & 1.7923 \\
LLM$_{\FS-10}$  & 0.4973 & 0.5160
                & 0.5214 & 0.4740
                & 0.6233 & 0.6114
                & 1.4735 & 1.4661
                & 5.9601 & 4.2810
                & 1.5178 & 1.4493 \\
LLM$_{\FS-30}$  & 0.6110 & 0.6354
                & 0.5164 & 0.5245
                & 0.6251 & 0.6276
                & 2.7207 & 2.3669
                & 6.7362 & 4.6829
                & 1.8060 & 1.4808 \\
LLM$_{\FS-50}$  & 0.5749 & 0.6027
                & 0.4572 & 0.4682
                & 0.5312 & 0.5843
                & 2.7465 & 2.5133
                & 6.3208 & 4.3760
                & 1.8499 & 1.3644 \\
\hline
\end{tabular}
}
\end{table*}

%% file: tables/table-knowledge-select.tex
\begin{table}[!ht]
\caption{
    Ablation study of \model on \bace, \esol and \lipo with variants of contextual knowledge retrieval. 
}
\label{table:influence_top_jump_random}
\centering
\small
\resizebox{1.\linewidth}{!}{
\begin{tabular}{l|c|c|c}
\hline
\hline
 & \textbf{\bace} & \textbf{\esol} & \textbf{\lipo} \\
\hline
 & ROC-AUC $\uparrow$ & RMSE $\downarrow$ & RMSE $\downarrow$ \\
\hline
GIN$^\model$& 0.8214 & 0.9137 & 0.6995 \\
w/ Jump     & 0.7799 & 1.0696 & 0.8744 \\
w/ Random   & 0.7868 & 1.1116 & 0.9027 \\
\hline
LM$^\model$ & 0.6915 & 1.3747 & 0.9468 \\
w/ Jump     & 0.6759 & 1.9781 & 1.4633 \\
w/ Random   & 0.6534 & 2.0615 & 1.9517 \\
\hline
\end{tabular}
}
\end{table}

%% file: pages/conclusion.tex
We have introduced a novel framework, \model, a training-free, lightweight, yet effective approach, harnessing the in-context learning capabilities of LLMs to improve the predictions of arbitrary ML models. 
Through this simple and versatile approach, we have demonstrated significant improvements over a number of ML models on different challenging tasks. 
As LLMs continue to improve in performance and in-context learning capabilities, \model stands to directly benefit from these advancements. 


%% file: pages/statement.tex
\section{Limitations and Ethic Statement}

\textbf{Limitations.}
While \model demonstrates simplicity and effectiveness in improving the predictions of an arbitrary ML model, our verification was mainly confined to structured molecular graph property prediction tasks and several text analysis tasks.
Further extensive empirical investigations across diverse domains are warranted to establish its generalisability.
Additionally, considering the purported enhanced ICL capabilities of GPT-4 on various benchmark tasks \cite{openaigpt4}, it is noteworthy that our findings (as discussed in Section~\ref{subsec:main_results} and illustrated in Table~\ref{table:influence_llms}) reveal GPT-4's underperformance compared to the GPT-3.5 model.
This discrepancy merits further exploration to elucidate the underlying reasons.
Moreover, while \model's prompt template could accommodate the insertion of molecule atom features and geometry structure descriptions, similar to prompt templates shown in Appendix~\ref{sec:appendix_additional_experiments}, limitations stemming from the LLM's input token constraints prevented their inclusion in the prompt in this paper.
We believe that with the rapid development of LLMs, some LLMs that allow much longer inputs will soon be available.
It would be interesting to investigate \model's effectiveness while including descriptions in the prompt. 
Lastly, while our approach incorporates contextual knowledge into the prompt, its utility is constrained by several factors, including limited flexibility. 
For example, further leveraging different techniques, \eg, RAG \cite{LPPP20} to involve more contextual knowledge into the LLM and advanced contextual knowledge retrieval~\cite{ZFT22} are also fruitful directions.
Further enhancements in this regard are warranted to maximise \model's effectiveness.

\smallskip\noindent
\textbf{Ethic Statement.} 
Our proposed framework, \model, is designed as a post-hoc \emph{corrector} aims at improving the prediction of an arbitrary ML model. 
However, given the emergent in-context learning ability within LLMs, which typically consist of billions of parameters, the accessibility of computational resources may inadvertently introduce disparities in the utilisation of these methods.
Research groups with limited access to computational resources will be handicapped, while resourceful groups will be able to investigate and advance the future directions of this research.
Throughout our work, we did not utilise any private or sensitive information.
However, it's essential to note that if any private information were to be inadvertently exposed to an LLM during internal pre-training and fine-tuning stages, \model does not offer any privacy filtration mechanism.
Therefore, there exists the potential for privacy concerns associated with the underlying model to manifest through the output provided by \model. 

%% file: pages/appendix.tex
\section{Illustration of Molecule Representations}

\begin{figure}[!ht]
\centering
\includegraphics[width=.9\linewidth]{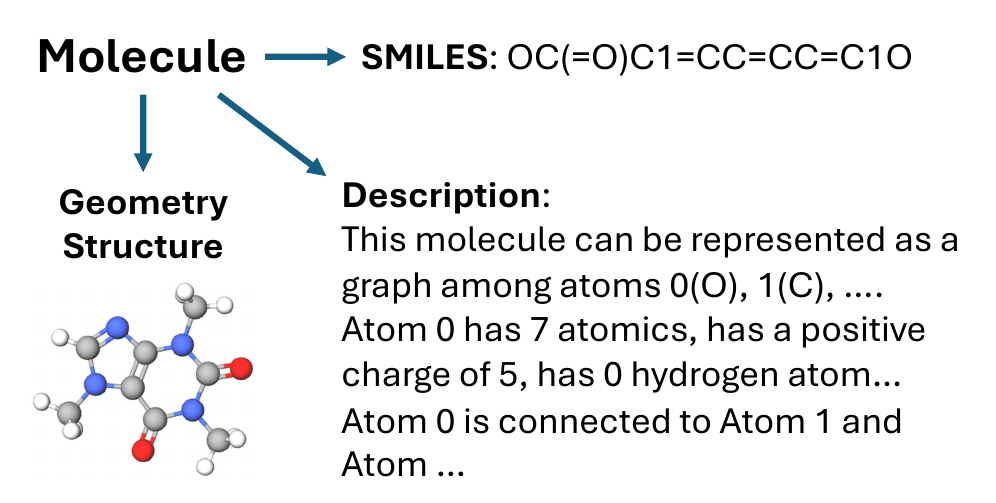}
\caption{
A molecule can be represented in different forms, \eg, SMILES string, text description and geometry structure. 
}
\label{fig:molecule_info}
\end{figure}

Molecules can be represented using various formats such as \textit{SMILES string}~\cite{W88} and \textit{geometry structures}~\cite{ZDLW24} (as shown in Figure~\ref{fig:molecule_info}). 
However, a notable limitation of existing LLMs is their reliance on unstructured text, rendering them unable to incorporate essential geometry structures as input~\cite{LLWL23,GDL23}. 
To overcome this limitation, \citet{FHP23} propose encoding the graph structure into text descriptions. 
In this paper, as depicted in Figure~\ref{fig:molecule_info}, we extend this concept by encoding both the molecule's atom features and graph structure into textual \textit{descriptions}.

\section{\model Prompt Templates for Regression Tasks}
\begin{figure}[!ht]
\centering
\includegraphics[width=1.\linewidth]{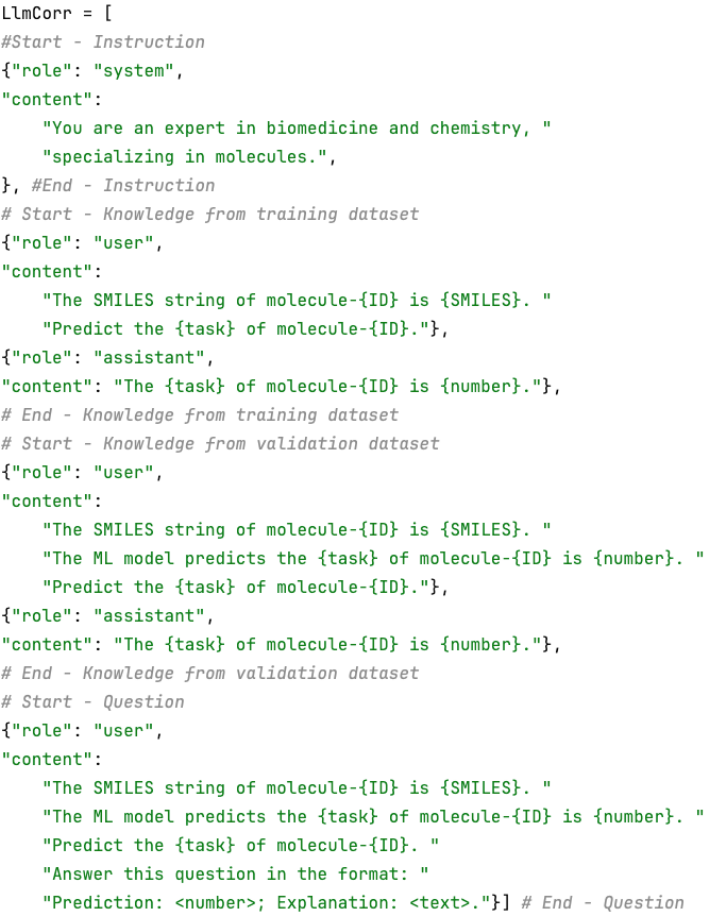}
\caption{
\model prompt template for regression tasks. 
Multiple contextual knowledge from training and validation datasets can be included by expanding the template. 
}
\label{fig:prompt_corrector_reg}
\end{figure}

\begin{figure}[!ht]
\centering
\includegraphics[width=1.\linewidth]{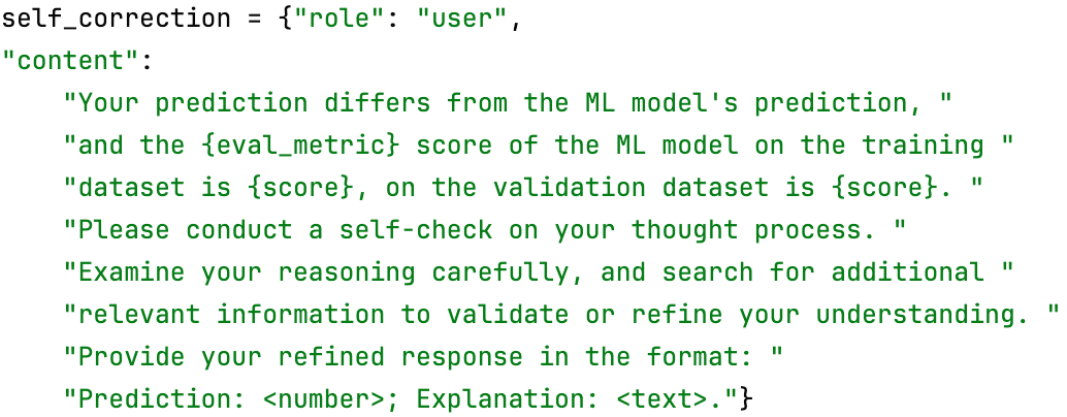}
\caption{
\textit{Self-correction} prompt template for regression tasks. 
}
\label{fig:prompt_self_correction_reg}
\end{figure}

Due to the page space limit, we only provide \model prompt templates in Section~\ref{sec:framework}. 
This section illustrates our \model prompt templates for regression tasks. 
Particularly, the proposed \model prompt template for regression tasks is illustrated in Figure~\ref{fig:prompt_corrector_reg}, which contains similar components as \model prompt template for classification tasks as discussed in Section~\ref{subsec:prompt_engineering}. 
The self-construction prompt for regression tasks is illustrated in Figure~\ref{fig:prompt_self_correction_reg}. 

\section{Algorithm}
\label{sec:appendix_algorithm}

\begin{algorithm}[!ht]
\SetAlgoLined
\KwIn{
Dataset $\mathcal{M} = \{ \mathcal{G}_{1}, \mathcal{G}_{2}, \dots, \mathcal{G}_{m}\}$, ML model $f_{ML}$, LLM $f_{LLM}$ 
}
\KwOut{
Refined predictions $\Tilde{\mathcal{Y}}$
}
    Complete training of $f_{ML}: \mathcal{M} \to \hat{\mathcal{Y}}$ by $\min_{\Theta} \sum_{i=1}^{n} \mathcal{L}(\hat{\mathcal{Y}}_{train}^{i}, \mathcal{Y}_{train}^{i})$ \; 
    Construct a contextual knowledge database $\mathcal{D} = \{\mathcal{M}_{train}, \mathcal{M}_{val}, \hat{\mathcal{Y}}_{val} \}$ \;
    \For{$\mathcal{G}_{u} \in \{ \mathcal{M}_{val} \cup \mathcal{M}_{test} \} $}{
        $\hat{y}_{u} = f_{ML}(\mathcal{G}_{u})$ \\
        $\mathcal{Q}_u = (\mathcal{G}_u, \hat{y}_u)$ \\
        Create a prompt $P_u$ using $\mathcal{Q}_u$ and retrieved contextual knowledge $\mathcal{D}_u \subset \mathcal{D}$ \\
        Query the LLM and contain the refined prediction $\Tilde{y}_u = f_{LLM}(P_u)$
    }
\caption{\model}
\label{alg:framework}
\end{algorithm}

We outline the process of \model in Algorithm~\ref{alg:framework}. 
Given a dataset $\mathcal{M}$, an ML model $f_{ML}$, a LLM $f_{LLM}$.
After completing the training of the ML model ($f_{ML}$) on the training set $\mathcal{M}_{train}$ (line 1), we construct a contextual knowledge database $\mathcal{D}$ by incorporating the dataset's label information and the ML model's prediction on the validation dataset $\mathcal{M}_{val}$ (line 2). 
Given a query data $\mathcal{G}_u$, we create a prompt $P_u$ using its primary prediction generated by $f_{ML}$ and relevant contextual knowledge $\mathcal{D}_u$ (line 3-6).
Finally, we query the LLM ($f_{LLM}$) to obtain the refined prediction $\Tilde{y}_{u}$ (line 7). 

\section{Dataset Description}
\label{sec:appendix_dataset}

\input{tables/table-dataset}

We consider six benchmark molecule property prediction datasets that are common within ML research, which are summarised in Table~\ref{table:dataset}. 
\begin{enumerate}[leftmargin=*]\itemsep0em
    \item \bace. The \bace dataset provides quantitative ($\mathrm{IC}_{50}$) and qualitative (binary label) binding results for a set of inhibitors of human b-secretase 1 (BACE-1). 
    All data are experimental values reported in the scientific literature over the past decade, some with detailed crystal structures available. 
    MoleculeNet~\cite{WRFGGPLP18} merged a collection of 1,522 compounds with their 2D structures and binary labels, built as a classification task. 
    \item \bbbp. The Blood–Brain Barrier Penetration (BBBP) dataset comes from scientific studies on the modelling and prediction of barrier permeability. 
    As a membrane separating circulating blood and brain extracellular fluid, the blood–brain barrier blocks most drugs, hormones and neurotransmitters. 
    Thus penetration of the barrier forms a long-standing issue in the development of drugs targeting the central nervous system. 
    This dataset includes binary labels for over 2,039 compounds on their permeability properties. 
    Scaffold splitting is also recommended for this well-defined target.
    \item \hiv. The HIV dataset was introduced by the Drug Therapeutics Program (DTP) AIDS Antiviral Screen, which tested the ability to inhibit HIV replication for 41,127 compounds. Screening results were evaluated and placed into three categories: confirmed inactive (CI), confirmed active (CA) and confirmed moderately active (CM). We further combine the latter two labels, making it a classification task between inactive (CI) and active (CA and CM). 
    As we are more interested in discovering new categories of HIV inhibitors, scaffold splitting is recommended for this dataset. 
    \item \esol. ESOL is a small dataset consisting of water solubility data for 1,128 compounds. 
    The dataset has been used to train models that estimate solubility directly from chemical structures (as encoded in SMILES strings). 
    Note that these structures don't include 3D coordinates, since solubility is a property of a molecule and not of its particular conformers.
    \item \freesolv. The Free Solvation Database (FreeSolv) provides experimental and calculated hydration-free energy of small molecules in water. 
    A subset of the compounds in the dataset is also used in the SAMPL blind prediction challenge.
    The calculated values are derived from alchemical free energy calculations using molecular dynamics simulations. 
    We include the experimental values in the benchmark collection and use calculated values for comparison.
    \item \lipo. Lipophilicity is an important feature of drug molecules that affects both membrane permeability and solubility. 
    This dataset, curated from the ChEMBL database~\cite{MGBCDFMMMN19}, provides experimental results of the octanol/water distribution coefficient (log D at pH 7.4) of 4200 compounds.
\end{enumerate}

\section{Implementation}
\label{sec:appendix_implementation}

\smallskip\noindent
\textbf{Implementation.}
We implement ML predictive models following their available official implementations.
For instance, we adopt the available code of variant GNN models on the OGB benchmark leaderboards, \eg, GCN\footnote{\url{https://github.com/snap-stanford/ogb/tree/master/examples/graphproppred/mol}}, GIN\footnote{\url{https://github.com/snap-stanford/ogb/tree/master/examples/graphproppred/mol}}, HIG~\footnote{\url{https://github.com/TencentYoutuResearch/HIG-GraphClassification}} and PAS~\footnote{\url{https://github.com/LARS-research/PAS-OGB}}. 
About DeBERTa, we adopt its official implementation~\footnote{\url{https://huggingface.co/microsoft/deberta-v3-base}} and incorporate it within the pipeline of TAPE~\footnote{\url{https://github.com/XiaoxinHe/TAPE}}. 
For the GPT-3.5 and GPT-4, we simply call the API provided by OpenAI with default hyper-parameter settings. 
For the Llama2, we adopt the Llama-2-13b version and call their official implementation on \url{https://huggingface.co}. 
We empirically tried with some combinations of recommended important hyper-parameters, \eg, temperature and top\_P, yet did not observe significant improvement. 
To realise the embedding-based information retrieval for \model, we adopt two capable embedding models ($f_{Emb}$) provided by OpenAI~\footnote{\url{https://platform.openai.com/docs/models/embeddings}}, \eg, \emph{text-embedding-3-large} and \emph{text-embedding-3-small}. 
In this work, we mainly adopt \emph{text-embedding-3-large} for better empirical performance. 
We perform careful discussions about the influence of different variants in Section~\ref{subsec:ablation_study}.

\section{Are LLMs Predictors?}
\label{sec:appendix_additional_experiments}

Following the thorough demonstration of \model's efficacy as a post-hoc corrector in Section~\ref{subsec:main_results}, a fundamental question emerged: does \model's remarkable performance stem from the LLM's ability to comprehend and rectify the ML model's predictions, or does it possess an inherent capability to predict molecule properties?
To answer this question, undertake another series of empirical investigations. 
Specifically, we devise \emph{predictor} prompts that task LLMs with directly predicting molecule properties, devoid of any information regarding the predictions made by the ML model.
In the following sections, we will present our designed prompts and demonstrate the experimental results. 

\begin{figure}[!ht]
\centering
\includegraphics[width=1.\linewidth]{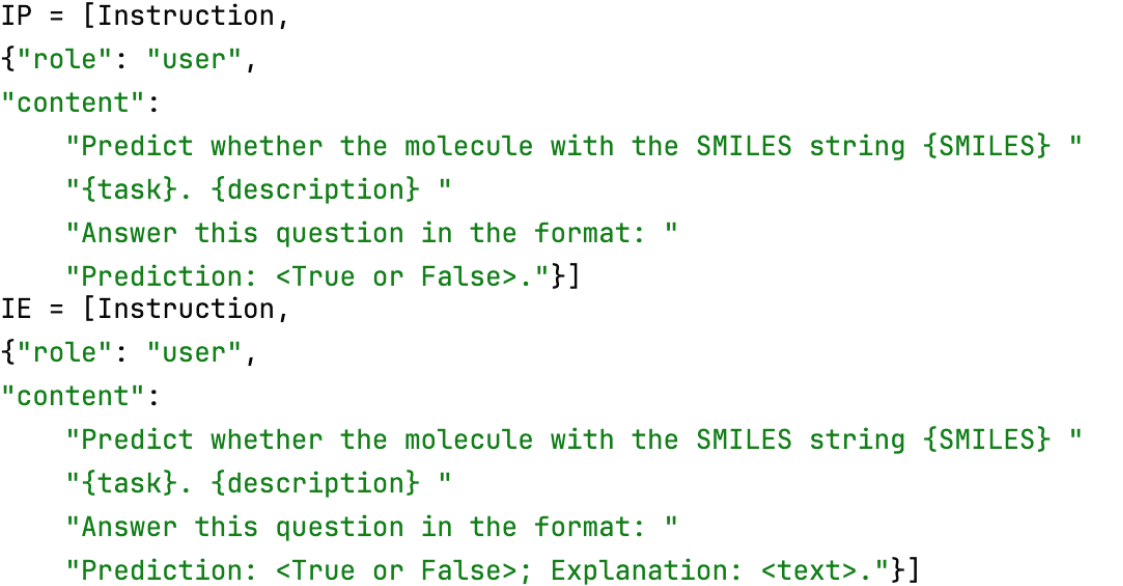}
\caption{
Zero-shot prompt templates for classification tasks. 
}
\label{fig:zero_shot_prompt}
\end{figure}

\begin{figure}[!ht]
\centering
\includegraphics[width=1.\linewidth]{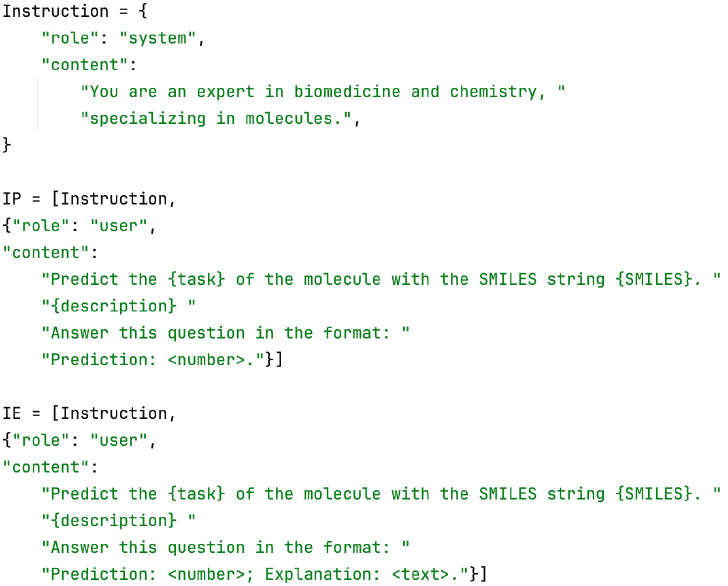}
\caption{
Zero-shot prompt templates for regression tasks. 
}
\label{fig:zero_shot_prompt_reg}
\end{figure}

\begin{figure}[!ht]
\centering
\includegraphics[width=1.\linewidth]{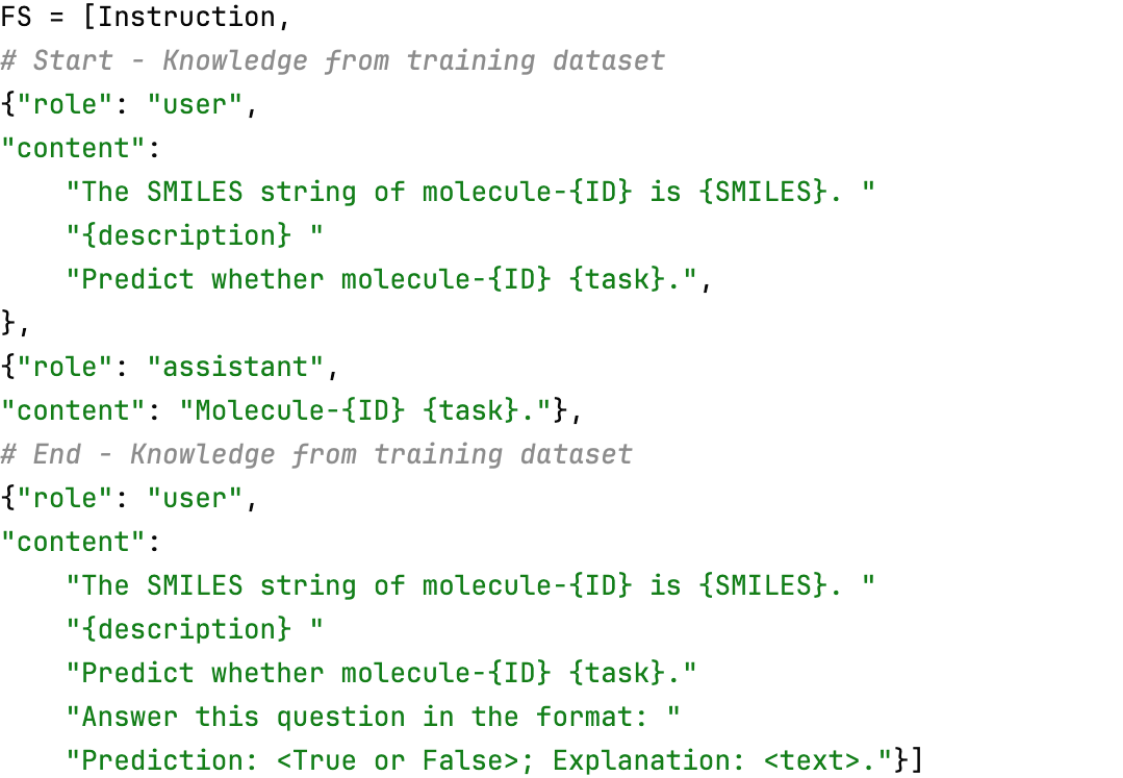}
\caption{
Few-shot prompt template for classification tasks. 
Multiple contextual knowledge can be included by expanding the template.
}
\label{fig:fs_prompt}
\end{figure}

\begin{figure}[!ht]
\centering
\includegraphics[width=1.\linewidth]{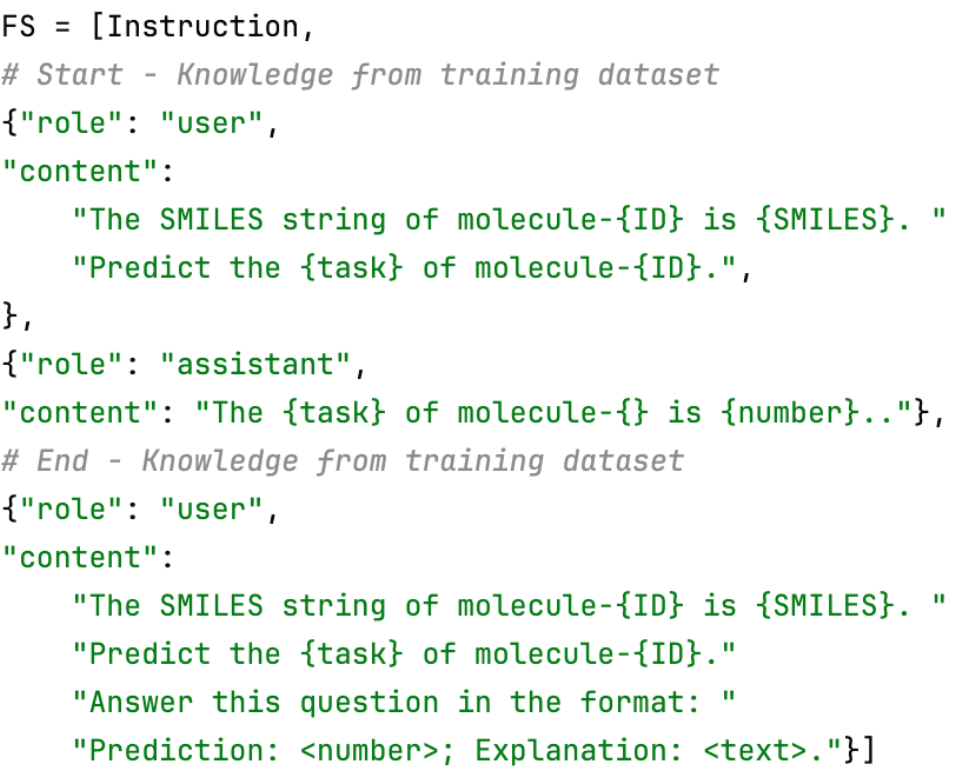}
\caption{
Few-shot prompt template for regression tasks. 
Multiple contextual knowledge can be included by expanding the template.
}
\label{fig:fs_prompt_reg}
\end{figure}

\subsection{Predictor Prompt Engineering}
\label{subsec:predictor_prompt:engineering}

\textbf{Zero-shot Prompting}.
The first set of prompts (\IP, \IE) simply provides the LLM with molecule and task descriptions and asks it to generate the desired output with a desired format without any prior training or knowledge on the task, as illustrated in Figure~\ref{fig:zero_shot_prompt} and Figure~\ref{fig:zero_shot_prompt_reg}. 
The only guidance we provide to the LLM is instruction, which tells about a little background context. 
Particularly, \IP only asks the LLM to provide predictions, while \IE further asks for explanations, which may ask the LLM to clarify the thought process in explanation generation and provide helpful evidence to help users understand the given prediction. 
In addition, if we fill out the \emph{description} of \IP and \IE, which derives \IPD and \IED prompts. 

\smallskip
\noindent
\textbf{Few-shot Prompting}.
The second kind of prompt (\FS) that we propose provides the LLM with a small number of examples of the task, along with the desired outputs~\cite{BMRS20}.
The model then learns from these examples to perform the task on new inputs. 
This approach can be categorised as a simple in-context learning (ICL) technique, 
An example prompt template is shown in Figure~\ref{fig:fs_prompt} and Figure~\ref{fig:fs_prompt_reg}. 
\FS-$k$ indicates $k$ contextual knowledge instances are included in the prompt. 
In this work, we do not discuss the \FSD prompts since the generated descriptions have tons of tokens, which will easily go over the LLM's input constraints. 

We note there are also some popular recent ICL techniques, \eg, Chain-of-thought (CoT)~\cite{WWSBIXCLZ22}, Tree-of-thought (ToT)~\cite{YYZSGCN23}, Graph-of-thought (GoT)~\cite{BBKG23} and Retrieval Augmented Generation (RaG)~\cite{LPPP20}, which are theoretically available to support complicated tasks and include large knowledge context. 
However, our initial experiments showed that methods, \eg, CoT, ToT and GoT, perform much worse for molecule property prediction tasks due to the significant difficulties in designing proper chain thoughts without solid expertise.
RaG implementations that we tested are unstable and slow with query, and they fall short of the relatively simpler \FS's performance. 
We argue the unqualified information retrieval system causes it, and we will discuss it in the future work discussion section. 

\subsection{Results - LLMs work as Predictors}
\label{subsec:appendix_additional_results}

From the results of Table~\ref{table:predictor_performance}, we can observe that the LLM can generate predictions about the molecule's property.
However, LLM's performances are not significantly competitive compared with the ML models' performance. 
Hence, we argue existing LLMs are not competitive predictors and employing LLMs as effective predictors is still an open challenge.

\section{\model for Other NLP Tasks}
\label{sec:appendix_broader_tasks}

In Section~\ref{sec:experimental_study}, we have illustrated the effectiveness of \model on various challenging molecular property prediction tasks. 
Our experimental results show that \model can consistently improve the performance of a number of models. 
In order to demonstrate the usability of \model in broader applications, we further conduct additional experiments on some NLP tasks. 

\smallskip\noindent
\textbf{Dataset.}
Particularly, we introduce two datasets:
\begin{enumerate}[leftmargin=*]\itemsep0em
    \item \twitter. The \twitter sentiment analysis dataset contains 6,940 tweets obtained by querying the Twitter API~\cite{DWTTZX14}. 
    Each tweet was manually annotated a sentiment labels (negative, neutral, positive). 
    The task is to predict the tweet's sentiment label based on its text. 
    We split datasets into training/validation/test (70\%, 10\%, 20\%) subsets for the experiments.
    
    \item \cora. The \cora dataset contains 2,708 academic publications~\cite{HBLPLH23}, which belong to one of seven categories (case-based, genetic algorithms, neural networks, probabilistic methods, reinforcement learning, rule learning, theory). 
    The task is to predict which category the publication belongs to based on its title and abstract. 
    We split datasets into training/validation/test (60\%, 20\%, 20\%) subsets for the experiments.
\end{enumerate}
We evaluate the model performance on these two datasets in terms of classification accuracy. 

\smallskip\noindent
\textbf{Methods.}
This section investigates the effectiveness of \model on NLP tasks, so we adopt NLP models of three categories: (1) one Language Model (LM), \ie, DeBERTa~\cite{HGC21}.
(2) One Large Language Model (LLM), \ie, GPT-3.5, which takes \IP and \FS prompts similar to the one shown in Figure~\ref{fig:zero_shot_prompt} and Figure~\ref{fig:fs_prompt}. 
(3) Our \model, that improves the predictions of DeBERTa. 
\model utilises a prompt similar to the one shown in Figure~\ref{fig:prompt_corrector}, by replacing \{SMILES\} and \{task\} with corresponding information. 
We adopt $k=10$ for \FS and \model in this experiment. 
Other settings are the same as the main experiments, discussed in Section~\ref{sec:experimental_study}. 

\input{tables/table-nlp-performance}

\smallskip\noindent
\textbf{Results.}
Examining the experimental results on two datasets in Table~\ref{table:nlp_performance}, it's evident that \model can consistently improve the LM's performance. 
It confirms the good usability of \model on various application tasks.
Meanwhile, we observe LLM exhibit competitive performance compared to the LM which fine-tune on the datasets. 
It is attributed to the LLM's qualified large-scale training datasets, which empower the LLM with an outstanding ability to understand different texts. 
Yet, compared to \model, our proposed framework still demonstrates significant priority over using LLMs as predictors. 

%% file: tables/table-dataset.tex
\begin{table*}[!ht]
\caption{
    Statistics summary of datasets used in our empirical study and splits from benchmark~\cite{WRFGGPLP18,HFZDRLCL20}. 
}
\label{table:dataset}
\centering
\resizebox{1.\linewidth}{!}{
\begin{tabular}{l|r|r|r|r|r|r|r}
\hline
\hline
\textbf{Dataset} & \textbf{\#Graphs} & \textbf{\specialcell{Avg. \\ \#Nodes}}  & \textbf{\specialcell{Avg. \\ \#Edges}} &  \textbf{\#Train} & \textbf{\#Valid} & \textbf{\#Test} & \textbf{Task Type} \\
\hline
\bace~\cite{WRFGGPLP18} & 1,513 & 34.1 & 73.7 & 1,210 & 151 & 152 & Binary class. \\
\bbbp~\cite{WRFGGPLP18} & 2,039 & 24.1 & 51.9 & 1,631 & 204 & 204 & Binary class. \\
\hiv~\cite{WRFGGPLP18,HFZDRLCL20} & 41,127 & 25.5 & 27.5 & 32,901 & 4,113 & 4,113 & Binary class. \\
\esol~\cite{WRFGGPLP18} & 1,128 & 13.3 & 27.4 & 902 & 113 & 113 & Regression \\
\freesolv~\cite{WRFGGPLP18} & 642 & 8.7 & 16.8 & 513 & 64 & 65 & Regression \\
\lipo~\cite{WRFGGPLP18} & 4,200 & 27.0 & 59.0 & 3,360 & 420 & 420 & Regression \\
\hline
\end{tabular}
}
\end{table*}

%% file: tables/table-nlp-performance.tex
\begin{table}[!ht]
\caption{
    Results on classic NLP tasks.  
}
\label{table:nlp_performance}
\centering
\begin{tabular}{l|c|c}
\hline
\hline
 & \textbf{\twitter} & \textbf{\cora} \\
\hline
 & \multicolumn{2}{c}{Accuracy $\uparrow$} \\
\hline
LM              & 0.7231 & 0.7626 \\
LLM$_{\IP}$     & 0.5493 & 0.6552 \\
LLM$_{\FS-10}$  & 0.6402 & 0.7123 \\
LM$^{\model}$   & 0.7420 & 0.7992 \\
\hline
\end{tabular}
\end{table}